\pdfoutput=1

\documentclass[11pt,dvipsnames]{article}

\usepackage[pagebackref=true]{hyperref}

\usepackage[final]{acl}

\usepackage{times}
\usepackage{latexsym}

\usepackage{footnote}
\usepackage{xcolor} 
\usepackage[normalem]{ulem}
\usepackage[T1]{fontenc}

\usepackage[utf8]{inputenc}

\usepackage{microtype}

\usepackage{inconsolata}

\usepackage{tikz}
\usetikzlibrary{arrows.meta,shapes,positioning,shadows.blur,trees}
\usepackage{fontawesome5} 
\usepackage{amsmath}
\usepackage{booktabs}
\usepackage[edges]{forest}
\usepackage{cleveref}
\usepackage{natbib}
\usepackage{amssymb} 
\usepackage{anyfontsize}
\usepackage{caption}
\usepackage{color}
\usepackage{enumitem}
\usepackage{float}
\usepackage{nicefrac}
\usepackage{rotating}
\usepackage{soul}
\usepackage{xcolor}
\usepackage{colortbl}
\usepackage{transparent}
\usepackage{wrapfig}
\usepackage{multicol}
\usepackage{multirow}
\usepackage{makecell}
\usepackage{array}
\usepackage{tabularx}
\usepackage{csquotes}
\usepackage{epigraph}
\usepackage[most]{tcolorbox} 
\usepackage{arydshln}
\usepackage{tablefootnote}
\usepackage{bbding}
\usepackage{pifont}
\newcommand{\cmark}{\ding{51}}%
\newcommand{\xmark}{\ding{55}}%

\crefformat{section}{(\S#2#1#3)}
\crefformat{subsection}{(\S#2#1#3)}
\crefformat{subsubsection}{(\S#2#1#3)}

\definecolor{lightred}{rgb}{0.749,0.196,0.447}
\definecolor{brown}{rgb}{0.65490196078,0.54901960784,0.48235294117}
\definecolor{greenn}{HTML}{32CD32}
\definecolor{DarkGreen}{rgb}{0,0.40,0}
\definecolor{FireBrick}{rgb}{0.698,0.133,0.133}
\definecolor{purple}{rgb}{0.5,0,0.5}

\title{Two Tales of Persona in LLMs:\\A Survey of Role-Playing and Personalization}

\author{
        Yu-Min Tseng\textsuperscript{\rm *}$^\dag$$^\ddag$\quad 
        Yu-Chao Huang\textsuperscript{\rm *}$^\dag$\quad 
        Teng-Yun Hsiao\textsuperscript{\rm *}$^\dag$\quad 
        Wei-Lin Chen\textsuperscript{\rm *}$^\dag$$^\star$\quad \\
    \textbf{
        Chao-Wei Huang$^\dag$\quad
        Yu Meng$^\star$\quad
        Yun-Nung Chen$^\dag$   
    } \vspace{5pt} \\
    $^\dag$National Taiwan University\quad
    $^\ddag$Academia Sinica\quad
    $^\star$University of Virginia\\
    \texttt{ymtseng@nlg.csie.ntu.edu.tw} \\
    \texttt{yumeng5@virginia.edu\quad y.v.chen@ieee.org}
    \vspace{5pt}\\
    \faGithub~ \url{https://github.com/MiuLab/PersonaLLM-Survey}
}
\begin{document}
\maketitle

\begin{abstract}
The concept of \textit{persona}, originally adopted in dialogue literature, has re-surged as a promising framework for tailoring large language models (LLMs) to specific context (\textit{e.g.}, personalized search, LLM-as-a-judge).
However, the growing research on leveraging persona in LLMs is relatively disorganized and lacks a systematic taxonomy.
To close the gap, we present a comprehensive survey to categorize the current state of the field.
We identify two lines of research, namely (1) \textit{LLM Role-Playing}, where personas are assigned to LLMs, and (2) \textit{LLM Personalization}, where LLMs take care of user personas.
Additionally, we introduce existing methods for LLM personality evaluation.
To the best of our knowledge, we present the first survey for role-playing and personalization in LLMs under the unified view of persona.
We continuously maintain a paper collection to foster future endeavors.
\begingroup\def\thefootnote{\rm *}\footnotetext{Equal contribution.}\endgroup
\end{abstract}

\begin{figure}[!ht]
    \centering
    \includegraphics[width=0.9\linewidth]{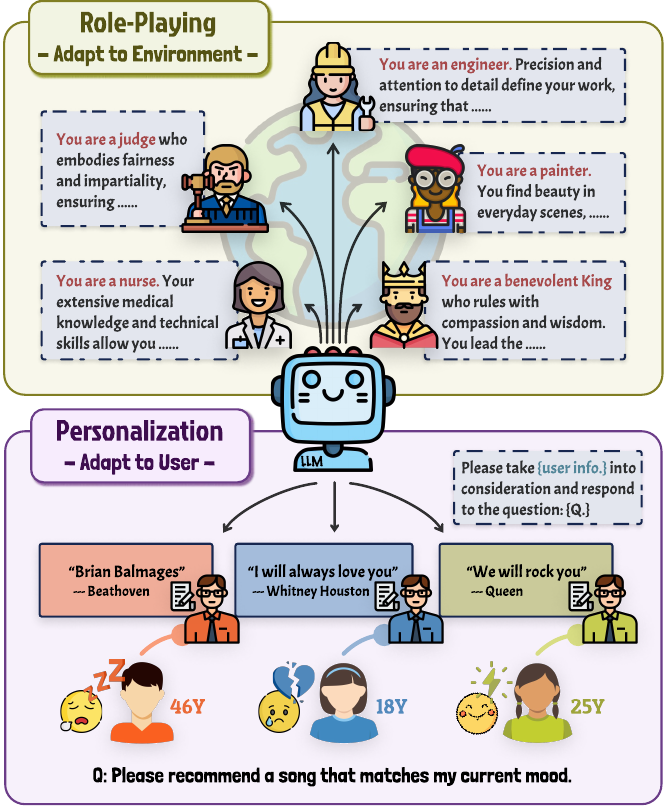}
    \caption{In \textit{Role-Playing}, LLMs act according to assigned personas (\textit{i.e.}, roles) under a defined environment. For example, given \textcolor{BrickRed}{\textit{role names}} with \textit{descriptions}, LLMs role-play in a social simulation game. For \textit{Personalization}, LLMs consider \textcolor{Cyan}{\textit{user personas}} to generate tailored responses to the same question. Dashed rectangles are prompts and solid rectangles are LLMs' responses.}
    \label{fig:overview}
\end{figure}

\section{Introduction}\label{sec:intro}
The striking capabilities of large language models (LLMs), exemplified by ChatGPT~\cite{chatgpt}, have significantly advanced the field of natural language processing (NLP;~\citealp{wei2023chainofthought,madaan2024self,shinn2024reflexion}).
Recently, in addition to using LLMs as NLP task solvers or general-purpose chatbots, the question of \textit{how to adapt~LLMs for specific context} has received great attention.
To this end, leveraging \textit{personas} has resurfaced as an ideal lens for adapting LLMs in target scenarios~\citep{chen2023large, chen2024persona}.
By incorporating personas, LLMs can generate more contextually appropriate responses, maximizing their utility and effectiveness for specific applications.
However, the growing literature on persona in the LLM era is relatively disorganized, lacking a unifying overview.
\definecolor{rootColor}{HTML}{FFFFFF} 
\definecolor{onodeColor}{HTML}{B3CDE3}
\definecolor{tnodeColor}{HTML}{FFF0C5}
\definecolor{edgeColor}{HTML}{000000}
\definecolor{textColor}{HTML}{000000}

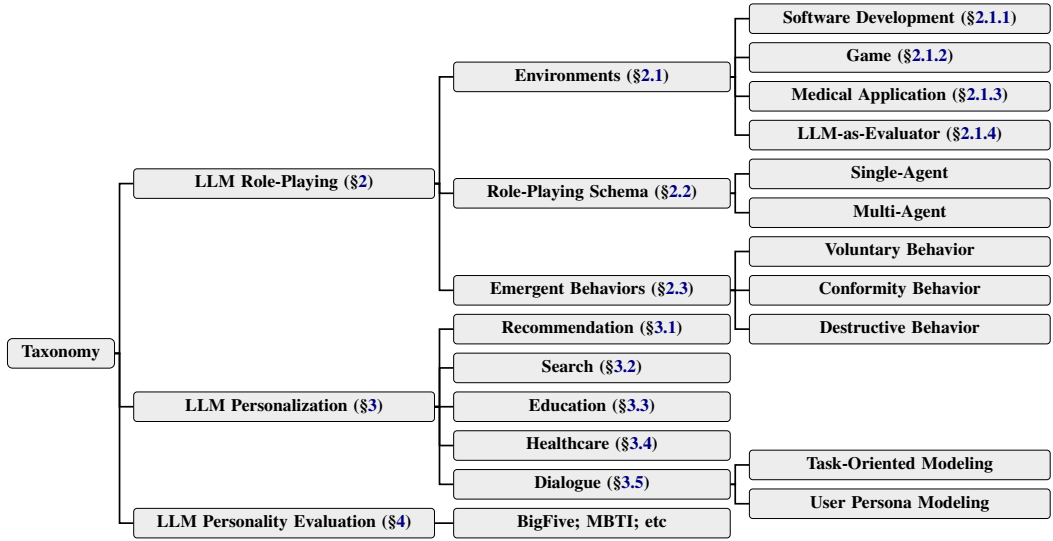
\begin{figure*}[t!]


\centering
\resizebox{\textwidth}{!}{%
\begin{forest}
  for tree={
    forked edges,
    draw=edgeColor,
    thick,
    font=\sffamily,
    fill=tnodeColor, 
    rectangle,
    rounded corners=4pt,
    text=textColor,
    blur shadow={shadow scale=0.95, shadow xshift=.5ex, shadow yshift=-.5ex, shadow opacity=0.25},
    edge={-, draw=edgeColor, line width=2pt}, 
    grow=0,
    child anchor=west,
    parent anchor=east,
    anchor=west,
    align=center,
    l sep+=0.3cm,
    s sep+=0.3cm,
  },
  root/.style=  {fill=Gray!15, font=\bfseries\huge, rounded corners=6pt, text width=4cm, /tikz/align=center, inner sep=6pt},
  tnode2/.style={fill=Gray!15, font=\bfseries\huge, text width=11cm, /tikz/align=center, inner sep=6pt},
  tnode3/.style={fill=Gray!15, font=\bfseries\huge, text width=9cm, /tikz/align=center, inner sep=6pt},
  tnode4/.style={fill=Gray!15, font=\bfseries\huge, text width=10cm, /tikz/align=center, inner sep=6pt},
  cnode4/.style={fill=Yellow!15, font=\LARGE, edge=dotted, text width=22.7cm, align=left},
  cnode5/.style={fill=Yellow!15, font=\LARGE, edge=dotted, text width=11.5cm, align=left},
  [Taxonomy, root
    [LLM Personality Evaluation~\Cref{sec:personality}, tnode2
        [Big Five; MBTI; etc, tnode3
            [~\citet{jiang2023personallm, sorokovikova-etal-2024-llms};\\
            ~\citet{pan2023llms, song2024identifying};
            ~PsychoBench~\citep{huang2023humanity}, cnode4]
        ]
    ]
    [LLM Personalization~\Cref{sec:personalized}, tnode2
        [Dialogue~\Cref{subsub:3-diag}, tnode3
            [User Persona Modeling, tnode4
                [~$\mathcal{P}^2$~\citep{liu-etal-2020-impress};
                ~\citet{kim2024commonsense};\\
                ~CLV~\citep{tang-etal-2023-enhancing-personalized}\\
                ~\textbf{-- Dataset --}\\
                ~PersonaChat~\citep{zhang-etal-2018-personalizing};\\
                ~ConvAI2 PersonaChat~\citep{dinan2019second}, cnode5] 
            ]
            [ToD Modeling, tnode4
                [~RefGPT~\citep{yang-etal-2023-refgpt};\\
                ~ProToD~\citep{hu2023enhancing};\\
                ~DSP~\citep{li2024guiding}\\
                ~\textbf{-- Dataset --}\\
                ~MutiWoz~\citep{budzianowski-etal-2018-multiwoz};\\
                ~SGD~\citep{rastogi2020towards}, cnode5] 
            ]
        ]
        [Healthcare~\Cref{subsub:3-healthcare}, tnode3
            [~openCHA~\citep{abbasian2024conversational};
            ~CHA~\citep{abbasian2024knowledgeinfused};\\
            ~MaLP~\citep{zhang2024llmbased};
            ~HealthLLM~\citep{jin2024healthllm}\\
            ~\textbf{-- Dataset --}\\
            ~MedDialog~\citep{zeng-etal-2020-meddialog};
            ~IMCS-21~\citep{chen2022benchmark};
            ~iCliniq~\citep{li2023chatdoctor}, cnode4],
        ]
        [Education~\Cref{subsub:3-education}, tnode3
            [~HumSum~\cite{shehata-etal-2023-enhancing};
            ~EduChat~\citep{dan2023educhat};
            ~\citet{park2024empowering};\\
            ~\citet{kasneci2023chatgpt};
            ~\citet{wang2024large}, cnode4]
        ]
        [Search~\Cref{subsub:3-search}, tnode3
            [~CoPS~\citep{zhou2024cognitive};
            ~\citet{spatharioti2023comparing};
            ~\citet{ziems2023large};\\
            ~\citet{sharma2024generative};
            ~\citet{baek2024knowledgeaugmented}, cnode4] 
        ]
        [Recommendation~\Cref{subsub:3-rec}, tnode3
            [~PALR~\citep{yang2023palr};
            ~\citet{Dai_2023};
            ~ONCE~\citep{liu2023once}\\
            ~\citet{christakopoulou2023large};
            ~BookGPT~\citep{zhiyuli2023bookgpt};\\
            ~\textbf{-- Dataset --}\\
            ~MovieLens~\citep{harper2015movielens}, cnode4]
        ]
    ]
    [LLM Role-Playing~\Cref{sec:role_play}, tnode2
        [Emergent Behaviors~\Cref{subsec:2-emergent}, tnode3
            [Destructive Behavior, tnode4
                [~CAMEL~\citep{li2024camel};\\
                ~\citet{deshpande2023toxicity};\\
                ~\citet{gupta2023bias}, cnode5]
            ]
            [Conformity Behavior, tnode4
                [~MedAgents~\citep{tang2023medagents};\\
                ~\citet{fu2023improving}, cnode5]
            ]
            [Voluntary Behavior, tnode4
                [~AgentVerse~\citep{chen2023agentverse};\\
                ~MetaGPT~\citep{hong2023metagpt}, cnode5]
            ]
        ]
        [Role-Playing Schema~\Cref{subsec:2-schema}, tnode3      
            [Multi-Agent, tnode4
                [~ChatDev~\citep{qian2023communicative};\\ ~MedAgents~\citep{tang2023medagents}, cnode5]
            ]
            [Single-Agent, tnode4 
                [~Voyager~\citep{wang2023voyager};\\
                ~MindAct~\citep{deng2024mind2web}, cnode5]
            ]
        ]
        [Environments~\Cref{subsec:2-env}, tnode3
            [LLM-as-Evaluator~\Cref{subsub:2-evaluator}, tnode4
                [~DRPE~\citep{wu2023large-drpe};\\
                ~ChatEval~\citep{chan2023chateval}, cnode5]
            ]
            [Medical Application~\Cref{subsub:2-medical}, tnode4
                [~\citet{wu2023large};\\
                ~MedAgents~\citep{tang2023medagents}, cnode5]
            ]
            [Game~\Cref{subsub:2-games}, tnode4
                [~Generative Agents~\citep{park2023generative};\\
                ~Humanoid Agent~\citep{wang2023humanoid};\\ 
                ~\citet{fu2023improving};\\
                ~Voyager~\citep{wang2023voyager}, cnode5]
            ]
            [Software Development~\Cref{subsub:2-software}, tnode4
                [~\citet{dong2023self};\\
                ~ChatDev~\citep{qian2023communicative};\\
                ~MetaGPT~\citep{hong2023metagpt}\\
                ~\textbf{-- Dataset --}\\
                ~SoftwareDev~\citep{hong2023metagpt};\\ 
                ~SRDD~\citep{qian2023communicative}, cnode5]
            ]
        ]
    ]
  ]
\end{forest}
}

\caption{The taxonomy of LLM role-playing and LLM personalization (representative works shown only).}
\label{fig:survey-struc-new}

\end{figure*}

In this paper, we aim to close the gap by offering a comprehensive survey and a systematic categorization of existing studies.
Specifically, we divide current research into two main streams, namely \textit{LLM Role-Playing} and \textit{LLM Personalization}, as illustrated in~\Cref{fig:overview}.
The primary distinction is that in role-playing, the persona belongs to the LLM, while in personalization, the persona belongs to the user.
%
Further, the literature on role-playing mainly focuses on the tasks (\textit{i.e.}, how LLMs with role-playing can achieve better performance). 
In contrast, the literature of personalization primarily focuses on the users (\textit{i.e.}, how to satisfy users’ expectations and meet their needs).
It is noteworthy that both of role-playing and personalization can be goals in the same scenario, but serve different purposes and are driven by different aspects.
The definitions are detailed below.
\begin{itemize}
\item \textbf{\textit{LLM Role-Playing}:} LLMs are tasked to play the assigned personas (\textit{i.e.,} roles) and act based on environmental feedback, adapting to the environment.
\item \textbf{\textit{LLM Personalization}:}
LLMs are tasked to take care of user personas (\textit{e.g.}, background information or historical behaviors) to meet individualized needs, adapting to distinct users.
\end{itemize}

To the best of our knowledge, we present the first survey for LLM role-playing and LLM personalization under the unified view of persona.
To foster future endeavors, we actively maintain a paper collection available to the research community.
We aim for this work to serve as both a valuable introduction for newcomers and a comprehensive resource for current researchers in the field.

Our taxonomy is illustrated in~\Cref{fig:survey-struc-new}.
We first introduce LLM role-playing~\Cref{sec:role_play}, followed by LLM personalization~\Cref{sec:personalized}.
Next, we provide an overview of evaluation methods~\Cref{sec:personality} assessing whether the personality of LLMs (\textit{e.g.}, personality traits or psychological behaviors) accurately aligns with expected personas after the adaptation (\textit{i.e.}, for role-playing LLMs that act according to assigned personas and personalized LLMs that fit user personas).
Lastly, we highlight current challenges and future directions~\Cref{sec:challenges}.
We hope that this taxonomy could serve as a useful guideline for researchers to easily target the tasks/scenarios of interests, and swiftly pinpoint their current position in the field.

\begin{figure*}[htp!]
    \centering
    \includegraphics[width=\linewidth]{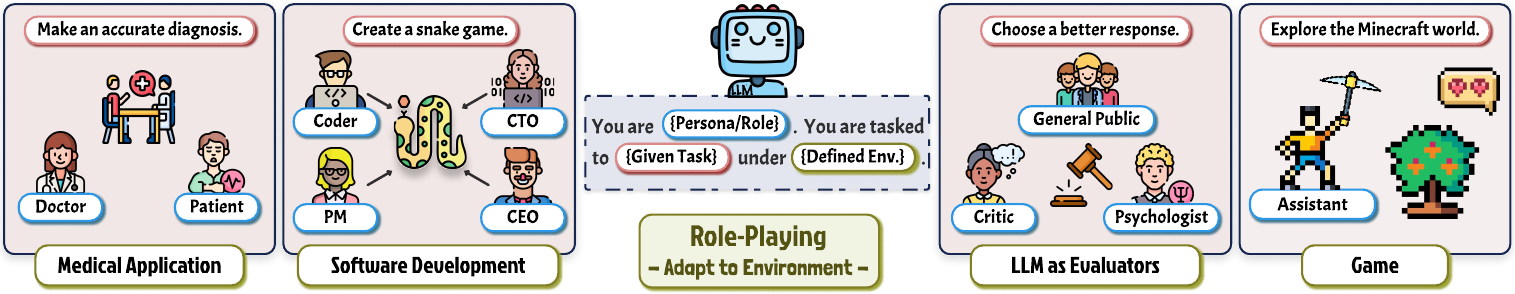}
    \caption{An illustration of four LLM role-playing environments: \textit{Software Development}~\Cref{subsub:2-software}, \textit{Game}~\Cref{subsub:2-games}, \textit{Medical Application}~\Cref{subsub:2-medical}, and \textit{LLM as Evaluators}~\Cref{subsub:2-evaluator}. For each environment, we provide a simple scenario with a task description (\textit{\textcolor{BrickRed}{red-bordered}}) and relevant personas (\textit{i.e.}, roles; \textit{\textcolor{CornflowerBlue}{blue-bordered}}). The dashed rectangle represents an example LLM role-playing prompt template. In addition to the above environments, past research also proposes general frameworks applicable to different environments~\Cref{subsub:2-general}.
    }
    \label{fig:overview_roleplay}
\end{figure*}

\section{LLM Role-Playing}\label{sec:role_play}
LLM-based language agents have demonstrated impressive abilities, such as planning, reflection, and tool-use recently~\cite{yao2022react,shinn2024reflexion,yao2024tree}.
The predominant approach of LLM role-playing is by coupling personas with language agents, specifically, by incorporating personas directly inside the prompt of language agents.
Such a training-free paradigm is particularly desirable due to its simplicity and effectiveness.

Language agents with role-playing elicit the corresponding parametric knowledge in LLMs to generate responses aligned with assigned personas (\textit{i.e.}, role), enabling them to adapt to various interactive environments.
LLM role-playing also extends to \textit{multi-agent} settings, where multiple language agents are equipped with diverse personas, cooperating and communicating with each other to solve complex tasks~\citep{guo2024large}.
For instance, in one of the first works of role-played LLMs, \citet{park2023generative} propose \textit{generative agents}, which engage in a social simulation environment by mimicking human behaviors according to names, ages, and personality traits specified in the prompts.

Following we introduce different environments and associated roles in which LLMs adapt to~\Cref{subsec:2-env}, interactions between LLMs within the environment~\Cref{subsec:2-schema}, and emergent behaviors stemming from their interactions~\Cref{subsec:2-emergent}.
\Cref{fig:overview_roleplay} provides an illustrative overview.
\subsection{Environments}\label{subsec:2-env}
\subsubsection{Software Development}\label{subsub:2-software}
For software development, the goal typically involves designing programs or coding projects.
For instance, \textit{``Create a snake game.''} or \textit{``Create a Python program to develop an interactive weather dashboard.''}~\citep{hong2023metagpt}.
Due to the complexity of these tasks, often too intricate to be completed correctly on the first attempt, existing research leverages approaches like the Waterfall model~\cite{petersen2009waterfall, bassil2012simulation} or Standardized Operating Procedures (SOPs)~\cite{belbin2022team, demarco2013peopleware} to break down the tasks into manageable sub-tasks.

Similar to real-world settings, LLMs role-play to operate as a company in a collaborative, multi-agent software development environment~\cite{qian2023communicative, hong2023metagpt, dong2023self}.
Different roles include Chief Technology Officer (CTO), Chief Product Officer (CPO), Chief Executive Officer (CEO), Product Managers, Engineers, Reviewers, and Testers.
By assigning specific roles, LLMs are capable of carrying out tasks in a step-by-step and accurate manner.

Recent work \cite{dong2023self} proposed one of the first self-collaboration frameworks that encompasses division of labor and collaboration among multiple LLM agents, each acting as a specialized \textit{``experts''} to address complex code generation tasks.
Following the Waterfall model, ChatDev~\citep{qian2023communicative} divides the development process into a four-phase pipeline: designing, coding, testing, and documenting and proposes \textit{Chat Chain} to decompose each phase into a sequence of atomic sub-tasks. 
Differing from the above work, MetaGPT~\citep{hong2023metagpt} require LLM agents to generate structured outputs instead of free-text, demonstrating a significant increase in the success rate of target code generation.

\subsubsection{Game}\label{subsub:2-games}
LLMs have been an effective backbone for agents in a variety of game environments, including Minecraft~\citep{wang2023voyager}, social simulation~\citep{park2023generative, wang2023humanoid}, and bargaining game~\citep{fu2023improving}.
In these environments, LLMs are tasked to role-play as a general assistant~\cite{wang2023voyager}, or characters related to the environment, such as buyers and sellers~\cite{fu2023improving}.
Gaming environments usually contain a wide range of information, including settings, utilizable tools, and nearby situations, which presents challenges for LLMs to memorize and respond.
Thus, retrieval-based memory stream approaches are a crucial component for the effectiveness of language agents role-playing in the game environments~\citep{park2023generative, wang2023voyager}.

\subsubsection{Medical Application}\label{subsub:2-medical}
In medical domain environments, \citet{wu2023large} propose DR-CoT prompting, the first approach to leverage LLM role-playing for diagnostic reasoning.
By mimicking doctors underlying thought processes, DR-CoT exhibits a striking improvement from standard prompting.
Then,~\citet{kwon2024large} extend such success to image-based diagnosis via knowledge distillation, addressing the application in real-world clinical settings.
Another work, MedAgent~\citep{tang2023medagents}, introduces a multi-agent collaboration framework into medical reasoning through five processes: expert gathering, analysis proposition, report summarization, collaborative consultation, and decision making, to mimic actual medical scenarios.

These studies assign medically relevant personas to LLMs, ranging from general roles like doctor and patient to specific ones such as neurology and psychiatry experts.
Their research demonstrates LLMs inherently possess medical knowledge~\citep{lievin2024can}, enabling performance enhancement via LLM role-playing successfully.

\subsubsection{LLM-as-Evaluator}\label{subsub:2-evaluator}
The concept of adopting strong LLMs as evaluators has become a de facto framework for evaluating LM alignment.
It is shown that LLMs are capable of assessing human-like values in model responses, and judgments made by LLMs could reflect a higher correlation with human ground-truth than traditional metrics~\citep{chiang2023can, wang2023aligning, lin-chen-2023-llm}.

Aiming for a greater similarity with human evaluation, roles in LLM-as-evaluator environments span a broad spectrum, representing various perspectives of human beings in society, such as the general public, the critic, and the news author.
In LLM-as-a-judge~\cite{zheng2023judging}, LLMs role-play an impartial judge and consider factors such as helpfulness, 
relevance, accuracy, depth, and creativity.
\citet{wu2023large-drpe} propose DRPE to assess the quality of summarization by assigning LLMs statically objective roles and dynamically subjective roles based on task settings.
Another work, ChatEval~\citep{chan2023chateval}, further adds discussion rounds within roles to improve the evaluation process, simulating a judge group in reality.

\subsection{Role-Playing Schema}\label{subsec:2-schema}
We categorize two schemas in LLM role-playing environments: \textit{single-agent} and \textit{multi-agent}.

\paragraph{Single-Agent}\label{subsub:2-single-agent}
We define the single-agent schema as: One agent is able to achieve its goal independently without assistance from others, though multiple agents may coexist in the same environment.

Single-agent schema is most common in game environments, where LLMs attend more to environmental information and feedback rather than collaboration.
For example, Voyager~\citep{wang2023voyager} agents, playing general assistant roles, are tasked to continuously explore the defined environment, acquire diverse skills, and make novel discoveries in Minecraft.
Despite the presence of multiple Voyager agents in Minecraft, each agent is capable of exploring the gaming world on its own.

\paragraph{Multi-Agent}\label{subsub:2-multi-agent}
We define the multi-agent schema as: Supports (\textit{e.g.}, collaborate and communicate) from other agents are necessary for one agent to achieve its goal.

Software development and medical applications are the primary environments for multi-agent schema.
Similar to real world, interaction within environments is crucial.
Representative works like AgentVerse~\citep{chen2023agentverse} and ChatDev~\citep{qian2023communicative} both propose multi-agent frameworks that exchange information and cooperate to accomplish their tasks efficiently.
Further, we identify two collaboration paradigms in the multi-agent schema~\citep{xi2023rise, guo2024large}: \textit{Cooperative} and \textit{Adversarial}.
The cooperative paradigm facilitates information sharing among agents, for example, several works use message pools to store each agent's current state and ongoing tasks~\citep{hong2023metagpt, tang2023medagents, chen2023agentverse}.
For the adversarial paradigm, including debate, competition, and criticism, enhances the decision-making process and seeks more advantages by adopting opposing perspectives~\citep{chan2023chateval, fu2023improving}.

\begin{figure*}[htp!]
    \centering
    \includegraphics[width=0.95\linewidth]{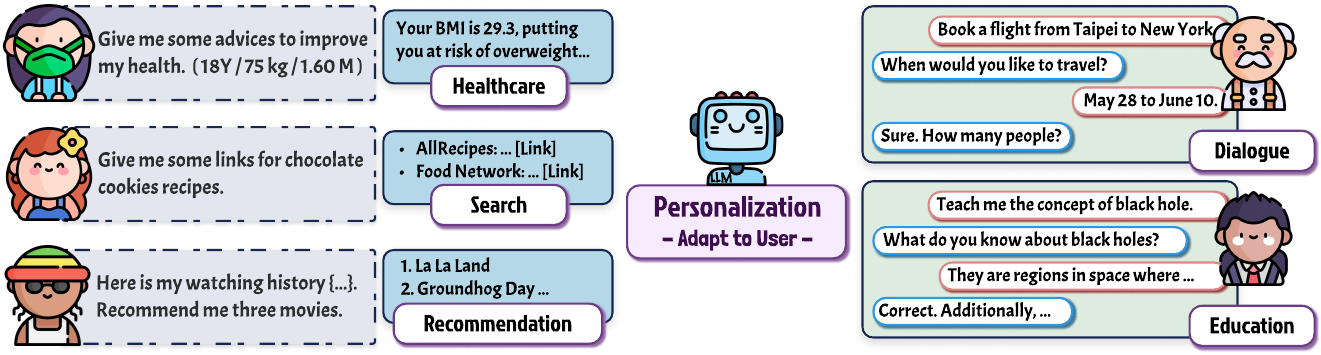}
    \caption{An illustration of five types of personalized LLMs:     \textit{Recommendation}~\Cref{subsub:3-rec}, \textit{Search}~\Cref{subsub:3-search},     \textit{Education}~\Cref{subsub:3-education}, \textit{Healthcare}~\Cref{subsub:3-healthcare}, and    \textit{Dialogue}~\Cref{subsub:3-diag}. On the left side, dashed rectangles are prompts, and solid rectangles are the responses of LLMs. On the right side, we depict multi-turn interactions between LLMs and users.
    }
    \label{fig:overview_personalized}
\end{figure*}

\subsection{Emergent Behaviors in Role-Playing}\label{subsec:2-emergent}
Under the multi-agent schema, different behaviors reflecting phenomena in human society (\textit{e.g.}, conformity and consensus reaching) emerge through LLM collaboration.
We introduce three collaborative behaviors following~\citet{chen2023agentverse}.

\paragraph{Voluntary Behavior}
Voluntary behaviors usually occur in the cooperative collaboration paradigm, where agents proactively assist their peers or inquire if there is anything they can help with to accomplish team goals.
In addition, they may contribute resources to others, such as unallocated time and possessed materials.
Through voluntary behaviors, LLMs enhance team efficiency and demonstrate cohesion and commitment within defined environments~\citep{chen2023agentverse, hong2023metagpt}.

\paragraph{Conformity Behavior}
Conformity behaviors occur in situations where an agent deviates from the team goal.
After receiving criticism and suggestions from others, the deviating agent then refines and adjusts its behavior or decisions to better cooperate with the team.
Through conformity behaviors, LLMs align with the mutual goal and pursue improved accuracy and completeness~\citep{tang2023medagents, fu2023improving}.

\paragraph{Destructive Behavior}
Occasionally, LLMs undertake various actions that lead to undesired and detrimental outcomes.
For instance, it may exhibit a \textit{Bad Mind} that seeks to control the world~\citep{li2024camel}.
Additionally, LLMs might display toxicity or reveal deep-seated stereotypical biases when equipping personas~\citep{deshpande2023toxicity, gupta2023bias}.
Such destructive behaviors raise safety and bias concerns of role-playing.

\section{LLM Personalization}\label{sec:personalized}

Prominent approaches for aligning LLMs to user intents typically leverage reinforcement learning from human feedback (RLHF), a process that infuses collective consciousness and biases into the model.
To enhance individual experience and preference, personalized LLMs consider user personas (\textit{e.g.,} individual information, historical behaviors) and cater to customized needs~\cite{chen2023first, personalize-2024-personalization}.
Following we introduce various personalized tasks with associated methods for achieving personalization.
\Cref{fig:overview_personalized} presents an illustrative overview of personalization tasks.

\subsection{Personalized Recommendation}\label{subsub:3-rec}
Recommendation systems aim to recommend items (\textit{e.g.}, books or movies) to users that match their preferences.
We compare existing research in~\Cref{tab:recommendation} and compile relevant datasets in~\Cref{tab:datasets_recommendation_search}.

Existing studies explore various prompting methods for using LLMs in recommendation systems. 
\citet{li2023personalized} develop a method for efficient incorporation of users' personal information.
\citet{li2023prompt} combine aspect extraction with aspect-based recommendations via LLMs prompt tuning.
\citet{chen2022personalized} generate personalized chit-chat to enhance recommendation.
Focusing on the framework design, \citet{yang2023palr} present a novel LLM fine-tuning recommendation system.
\citet{chu2023leveraging} merge different recommendation systems to effectively integrating the commonsense and reasoning abilities of LLMs.
\citet{hu2024enhancing} propose a sequential recommendation framework to preserve fine-grained item textual information.

A lot of works have focused on the zero-shot setting, leveraging the powerful out-of-the-box capabilities of LLMs.
\citet{wang2023zeroshot} adopt a three-step prompting pipeline to achieve better zero-shot next-item recommendation.
\citet{hou2024large} propose a zero-shot sequential recommendation system via in-context learning.
\citet{zhang2023recommendation} enhance user-friendliness by allowing users to freely interact with the system and receive more precise recommendations through natural language instructions.
For generalizability, \citet{wang2024recmind} highlight that current recommendation systems mostly focus on specific tasks and lack the ability to generalize to new tasks. They propose an LLM-powered agent for general recommendation purposes.
Although LLM-based personalized search systems present a more convenient and simple solution for information search, ensuring the accountability and trustworthiness of the synthesized results still requires further development~\cite{li2024survey}.

\subsection{Personalized Search}\label{subsub:3-search}
Compared to traditional search systems that provide a list of hard-to-organize relevant results and are limited to simple queries, personalized search systems enable understanding of complex queries and past interactions to infer user preferences, synthesizing information from multiple sources and presenting it in a cohesive, natural language form.

\citet{spatharioti2023comparing} demonstrate that LLM-based search systems improve users' performance in certain situations.
\citet{ziems2023large} suggest that LLMs act as built-in search engines given few-shot demonstrations.
Specifically, LLMs can generate correct web URLs for corresponding documents.
Building upon \citet{zhou2021group}, \citet{zhou2024cognitive} present a strategy to combine the cognitive memory mechanism with LLMs for personalized search, enabling LLMs to efficiently retrieve memory.
Some works also leverage search engine results to enhance LLM personalization~\cite{baek2024knowledgeaugmented, salemi2024towards}.
Empirically,~\citet{sharma2024generative} conduct experiments to investigate how LLM-powered search systems could lead to opinion polarization.

\subsection{Personalized Education}\label{subsub:3-education}
The capability of LLMs can be utilized in a variety of ways to facilitate personalized education.
For example, LLMs can provide detailed, step-by-step explanations in the Socratic teaching style~\citep{hao2024llm}, answer questions on technical and complicated subjects~\citep{arefeen2023leancontext}, and automatically summarize lectures to enhance learning experience~\citep{gonzalez-etal-2023-automatically}.

Personalized LLMs have the potential to create a more inclusive and equitable educational ecosystem, obviating the need for individuals to pay disproportionate fees.
Recent works have illustrated various opportunities and visions for integrating LLMs into educational environments.
These applications range from personalized learning and teaching assistance to homework assessment and feedback~\cite{kasneci2023chatgpt, wang2024large, jeon2023large, huber2024leveraging}.

For example, \textsc{EduChat}~\cite{dan2023educhat} pre-trained models on an educational corpus to establish a foundational knowledge base, and subsequently fine-tune models on personalized tasks such as essay assessment, Socratic teaching, or emotional support.
\textsc{HumSum}~\cite{shehata-etal-2023-enhancing} summarize personalized lecture transcripts from diverse scenarios, considering factors such as length, depth, tone, and complexity.
This is followed by prompt tuning to modify the summary based on the personalization options given by users.
\citet{park2024empowering} incorporate the student's affective state, cognitive state, and learning style into the prompt to create a personalized conversation-based tutoring system.

\subsection{Personalized Healthcare}\label{subsub:3-healthcare}
LLMs have exhibited expert-level capabilities in a range of general biomedical tasks, with the potential to integrate into people's everyday lives~\cite{cohan-etal-2020-specter,milne2020effectiveness,singhal2023large,saab2024capabilities,abbasian2024foundation}.

Towards personalized healthcare assistant,~\citet{abbasian2024conversational} propose \textsc{openCHA}, an LLM agentic framework that integrates external data and personalized health data to address personalized medical problems.
Following \textsc{openCHA}, \citet{abbasian2024knowledgeinfused} infuse domain-specific knowledge to effectively utilize health data, knowledge bases, and analytical tools for diabetes-related questions.
\textsc{MaLP}~\cite{zhang2024llmbased} combine parameter-efficient fine-tuning (PEFT) with a memory retrieval module to generate personalized medical responses.
Other frameworks such as \textsc{HealthLLM}~\cite{jin2024healthllm} combine LlamaIndex~\cite{Liu_LlamaIndex_2022} to make diagnosis predictions, and is capable of generating personalized medical advice based on symptom descriptions provided by users.
Moreover, LLMs also show great potential for psychotherapy~\cite{stade2024large, chen2023llm, xu2024mental}.

\subsection{Personalized Dialogue Generation}\label{subsub:3-diag}
Depending on the goals, dialogue generation tasks can be categorized into: (1) Task-oriented dialogue modeling (ToD modeling) and (2) User persona modeling.
Following we discuss ToD modeling and User persona.
We also organize various datasets for dialogue generation in~\Cref{tab:diag_data}.

\paragraph{ToD Modeling}
ToD modeling guides users in completing specific tasks, such as hotel bookings or restaurant reservations, through multiple interactive steps.
See an example in \Cref{tab:dialogue_example}.

\citet{hudecek-dusek-2023-large} leverage instruction-tuned LLMs and employ in-context learning for retrieval, and state tracking.
Focusing on factuality, \textsc{RefGPT}~\cite{yang-etal-2023-refgpt} generate truthful responses by augmenting the dialogue history with reliable sources and use prompts to guide LLM according to predefined dialogue settings.
\citet{li2024guiding, hu2023enhancing} explore prompt augmentations; on the other hand, \textsc{DSP}~\cite{li2024guiding} train a small policy model to generate hints and guide LLMs in completing tasks.
A lot of works used LLMs to generate multi-turn dialogue as training datasets~\citep{yang-etal-2023-refgpt,huryn-etal-2022-automatic,xu-etal-2023-baize}.
Further, personalized dialogues have been applied in procedural content generation for customized dialogue generation in video games~\cite{ashby2023personalized}.

\paragraph{User Persona Modeling}\label{sec:persona_modeling}
User persona modeling detects the user persona based on dialogue history and generates customized responses tailored for each user.
See an example in \Cref{tab:dialogue_persona}.

\textsc{CoBert}~\cite{zhong-etal-2020-towards} proposed persona-based empathetic conversations using BERT with a two-hop co-attention mechanism~\cite{lu2017hierarchical} to refine embeddings and identify the most relevant response given the context and persona information.
\citet{Song_Zhang_Hu_Liu_2020} utilized natural language inference (NLI) as an RL task with response persona as the reward to generate persona-consistent dialogue.
\citet{liu-etal-2020-impress} proposed $\mathcal{P}^2$, a mutual persona perception model, and employ supervised training and self-play fine-tuning in the training process.
\citet{tang-etal-2023-enhancing-personalized} combined sparse persona descriptions, dense persona descriptions, and dialogue history to generate personalized responses.

\section{LLM Personality Evaluation}\label{sec:personality}
In the previous sections, we summarize the current progress in LLM role-playing and LLM personalization.
Equally important is the evaluation of whether the personality of LLMs accurately reflects the intended persona after the adaptation (\textit{i.e.}, for role-playing LLMs that act based on designated personas and personalized LLMs tailored to individualized personas).

A line of works has carried out the evaluation leveraging human personality assessments, including Big Five~\cite{jiang2023personallm, sorokovikova-etal-2024-llms} and MBTI~\cite{pan2023llms, song2024identifying}.
For example,~\citet{sorokovikova-etal-2024-llms, jiang2024evaluating} quantitatively evaluate LLM personality based on the Big Five Personality Inventory (BFI) test and story writing test.
In the BFI evaluation, LLMs often can reflect their intended persona accurately. 
Moreover, their personas often influence their linguistic style and personality consistency \cite{frisch-giulianelli-2024-llm, jiang2023personallm}.
While most works focus solely on either semantic accuracy or personality consistency, \citet{harrison-etal-2019-maximizing} further explore controlling the two aspects simultaneously.

\citet{jiang2024evaluating} introduce Machine Personality Inventory (MPI) for evaluating LLMs' personality traits. 
They use Big Five Personality Factors to evaluate each personality trait consisting of a series of descriptions and a set of options and statistically measure each trait.
By comparing with human evaluation, they find that the internal consistency correlates with model capabilities.
On the other hand,~\citet{pan2023llms} evaluate LLMs with the MBTI test to assess whether LLMs possess human-like personalities, and conclude that different LLMs have different MBTI types, which are often attributable to their training corpus.
Moreover, they find that simply modifying the prompts is unlikely to change the MBTI type of LLMs.

Another work by~\citet{wang2024incharacter} evaluate the personality fidelity of role-playing LLMs via personality test interviewing, and ask LLM to rate the score of each personality dimension according to the interview.
Their results suggest that LLMs' demonstrated personalities align well with the assigned character personas.
However, whether the aforementioned human psychometric tests are directly transferable to be applied to LLMs remains an open question~\cite{dorner2023personality}.

\section{Challenges and Future Directions}\label{sec:challenges}

\subsection{Towards a General Framework}\label{subsub:2-general}
Despite the effectiveness of various role-playing frameworks, they are mostly task dependent and heavily rely on human-crafted personas.
Both require prior knowledge and deep understanding of the tasks~\citep{chen2023agentverse}.
Consequently, enhancing the generalizability of the framework and employing automatic prompt engineering is a fruitful directions~\citep{li2024camel, wang2023unleashing}.

To this end, \citet{li2024camel} propose a novel task-independent framework that allows agents to collaborate autonomously, but is limited to two roles and still requires human assigned personas.
Subsequently, \citet{wang2023unleashing} introduce methods for LLMs to automatically identify personas based on given problems.
Another work by~\citet{chen2023agentverse} also enable LLMs to dynamically adjust the personas.
However, they require prior knowledge of the intended tasks and pre-defined configuration (\textit{e.g.}, the number of agents).

\subsection{Long-Context Personas}
\citet{richardson2023integrating} note that incorporating user history data into the prompt for personalizing LLMs could lead to input exceeding context length as well as increased inference costs.
Leveraging retrieval-based methods may have the problem of potential information loss.
Some works have proposed to summarize user profiles, design long-term memory mechanisms focusing on user portrait, pre-storing user information, or ways to efficiently represent for retrieval augmentation~\cite{richardson2023integrating,Zhong_Guo_Gao_Ye_Wang_2024,zhang2024personalized,sun2024personadb}.
However, retrieval augmentation might be underperforming due to unrelated or noisy prompts~\cite{tan2024democratizing}.
How to better store, encode, and integrate long-context personas in LLMs requires further investigation.

\subsection{Lack of Datasets and Benchmarks}
For LLM role-playing, several tasks lack suitable datasets with specific formats~\cite{ahn2024timechara} and environmental information (\textit{e.g.}, game environments require information about configurations and tools).
For personalized dialogue generation, user persona modeling lacks contradictory persona datasets and multimodal persona datasets that would more accurately represent real human behaviors~\citep{kim2024commonsense, ahn2023mpchat}.
Furthermore, LLM personalization faces a scarcity of high-quality personal data for model development due to privacy concerns, hindering a thorough evaluation of personalization methods.
In addition, existing benchmarks for both LLM role-playing and personalization are relatively limited, lacking comprehensive evaluations across various dimensions~\citep{chang2023survey, samuel2024personagym}. 
Therefore, expanding datasets and benchmarks for specialized environments and personal information under privacy protection is an important next step.

\subsection{Bias}
While a large number of studies focus on enhancing end-task performance, fewer works explore the biases induced by role-playing and personalization in LLMs.
In this context, \citet{gupta2023bias}, as one of the first studies, highlight the deep-seated stereotypical biases found in LLMs assigned with socio-demographic personas.
Additionally, \citet{zhao2024bias} find that applying role-play often increases
the overall likelihood of generating stereotypical and harmful outputs.
For personalized LLM recommendation systems, biases can be observed due to item popularity or item positions in the prompts~\cite{hou2024large}.
Empirically, \citet{dorner2023personality} also reveal the presence of \textit{agree bias} in LLMs -- a tendency to agree with both true and false content, regardless of the actual facts.
In sum, there exists ample room for investigating and mitigating different classes of biases in the context of LLM role-playing and personalization.

\subsection{Safety and Privacy}
Past research has shown safety issues in LLM role-playing and personalization.
\citet{jin2024guard} and~\citet{shah2023scalable} successfully manipulate LLMs to perform jailbreak collaboratively.
\citet{deshpande2023toxicity} also show that assigning personas to LLMs aid in jailbreaking.
Negative behaviors in LLM role-playing are also demonstrated by \citet{chen2023agentverse} and~\citet{li2024camel}.
Further, \citet{deshpande2023toxicity} find that LLMs consistently exhibit toxicity in a range of topics when assigned personas, and ~\citet{vijjini2024exploring} show that LLMs suffer from \textit{personalization bias} when they are personalized for the user’s demographic.
These works demonstrate the discovery of unsafe problems, indicating an urgent need and more efforts to prevent potential exploits.

Since LLM personalization heavily relies on user personas, including personal information and historical behaviors, ensuring privacy is especially crucial.
Recently,~\citet{Wang2024PandorasWI} discover that using the membership inference attack can leak personal information, raising concerns about encoding personal data into models.
Although existing research provides methods to address this personal information leakage~\citep{lukas2023analyzing, Gambarelli_2023, huang-etal-2022-large, chen2023language}, the risks remain in need of more effort and attention from the research community.

\section{Broader Implications}
As LLM personalization continues to advance in education domains, individuals could easily access personalized educational contents, lecture materials, and receive affordable tutoring, largely benefiting minority groups with limited resources.
However, the concern of polarizing trends might arise, where the privileged group enjoys private tutors and underrepresented individuals only have access to LLM-powered supports~\cite{li2023adapting}.
Also, personalized LLMs for healthcare could potentially be widely integrated into clinical scenarios, mental health assessments, or prescribed therapeutic treatments in the near future, where critical questions such as legal considerations of the liability associated with these personalized systems needs careful considerations~\cite{ferguson2010towards}.

As discussed in~\Cref{sec:personality}, though methods for LLM personality evaluation have been proposed, there still lacks a unifying understanding of how to quantify personality in LLMs~\cite{fang-etal-2023-text}.
\citet{song2024identifying, jiang2024evaluating} also show that LLMs sometimes do not hold consistent personalities.
It is crucial to continuously explore new measurements for reliable assessment of personality and psychological traits in LLMs, considering that in the future they might take on more advanced roles and capabilities in society.

\section{Conclusion}\label{sec:conclusion}

Leveraging personas, LLMs can generate tailored responses and effectively adapt to a wide range of scenarios.
In this survey paper, we summarize two lines of work -- role-playing and personalization -- for research of personas in the era of LLMs.
We also present various evaluation methods for LLM personality.
Lastly, we highlight current challenges and promising future
directions.
We hope our extensive survey and resources serve as an introductory guide for beginners to the field and a practical roadmap to foster future endeavors.

\section*{Limitations}
For the evaluation metric, as the literature, even within the same scope, addresses various sub-tasks and employs different corresponding evaluation metrics, or proposes their own ones (e.g., persona accuracy, task success rate, combined inform and success rate). 
This largely increase the difficulty to establish a suitable/fair standard for comparison. 
Also, some scenarios may require multiple metrics to determine overall performance~\citep{samuel2024personagym}. 
For instance, we might need to assess the fluency, empathy, and safety of personalized LLMs. 
Consequently, we do not include a comprehensive evaluation comparison in the paper. 
Instead, we provide the solid taxonomy, content, and future directions that could serve as both a valuable introduction for newcomers and a comprehensive resource for current researchers in the field.

\section*{Acknowledgements}
We thank Yu-Ching Hsu and Jia-Yin Foo from National Taiwan University for their assistance and discussion.
This work was financially supported by the National Science and Technology Council (NSTC) in Taiwan, under Grants 112-2223-E-002-012-MY5 and 111-2222-E-002-013-MY3, and from the Featured Area Research Center Program within the framework of the Higher Education Sprout Project by the Ministry of Education (113L900901/113L900902/113L900903).

\bibliography{custom,anthology}

\appendix
\section{Web}\label{sec:2-web}
Prior works also investigate adapting LLM-based language agents to solve tasks in web environments.
However, they typically achieve this via task-independent instructions rather than specific role-playing.
Here we provide relevant research for leveraging LLMs in web environment.

In this environment, LLMs operate web navigation autonomously, performing actions such as clicking items, capturing contents, and searching from external knowledge on the web, without a specific persona assigned.
Certainly, web tasks involve two key components: \textit{HTML understanding} and \textit{visual grounding}, which are highly related to the effectiveness of web agents~\citep{zheng2024gpt, koh2024visualwebarena}.
Meanwhile, a stream of works, compiled in~\Cref{tab:web-bench}, proposes several benchmarks to assess web agents in diverse aspects.

\paragraph{HTML Understanding.} 
\citet{kim2024language} showcase that the ability of HTML understanding is inherent in LLMs with the Recursive Criticism and Improvement (RCI) prompting method.
However, due to the special formats and long context elements of HTML which are hard for LLMs to process and respond accurately, most research enhances this capability through fine-tuning methods~\citep{gur2022understanding, gur2023real, deng2024mind2web}.

\paragraph{Visual Grounding.}
Another line of research focuses on the visual grounding aspect of HTML understanding, which directly operates on rendered webpages instead of the HTML source code.
Some literature proposes web agent frameworks, such as CogAgent~\citep{hong2023cogagent} and SeeClick~\citep{cheng2024seeclick}, leveraging Large Multi-modal Models (LMMs)~\citep{achiam2023gpt, team2023gemini}.
With additional information from webpage screenshots, LMMs usually outperform text-based LLMs~\cite{zheng2024gpt}.

\begin{table*}[htbp]
  \footnotesize
  \centering
  \resizebox{0.9\textwidth}{!}{
    \begin{tabular}{lcccccc}
    
    \toprule
    \multirow{2}[2]{*}{\textbf{Benchmark}} & 
    \multirow{2}[2]{*}{\textbf{\#Instances}} & 
    \multirow{2}[2]{*}{\textbf{\#Domains}} & 
    \textbf{Realistic} & \textbf{Dynamic} & \textbf{Visual} & \multirow{2}[2]{*}{\textbf{Assessment}} \\
          &
          &
          &
    \textbf{Env.} &
    \textbf{Interaction} &
    \textbf{Needed} &  \\
    
    \midrule
    WebShop~\citep{yao2022webshop}    & 12,087 & 1 & \textcolor{BrickRed}{\xmark} & \textcolor{Green}{\cmark} & \textcolor{BrickRed}{\xmark}     & End-to-end \\
    Mind2Web~\citep{deng2024mind2web} & 2,350  & 5 & \textcolor{Green}{\cmark} & \textcolor{BrickRed}{\xmark} & \textcolor{BrickRed}{\xmark} & End-to-end \\
    WebArena~\citep{zhou2023webarena} & 812    & 4 & \textcolor{Green}{\cmark} & \textcolor{Green}{\cmark} & \textcolor{BrickRed}{\xmark}     & End-to-end \\
    VisualWebArena~\citep{koh2024visualwebarena} & 910    & 3 & \textcolor{Green}{\cmark} & \textcolor{Green}{\cmark} & \textcolor{Green}{\cmark} & End-to-end \\
    VisualWebBench~\citep{liu2024visualwebbench} & 1,500  & 12 & \textcolor{Green}{\cmark} & \textcolor{BrickRed}{\xmark} & \textcolor{Green}{\cmark} & Fine-grained \\
    
    \bottomrule
    \end{tabular}
  }
  \caption{Comparison between recent benchmarks in the web environment. \textit{Realistic Env.} denotes whether the benchmark's environments are based on actual web pages or realistic web navigation simulations. \textit{Dynamic Interaction} indicates whether the benchmark supports dynamic interactions rather than remaining in static states. \textit{Visual Needed} denotes whether the benchmark involves visually grounded tasks. \textit{Assessment} refers to the types of assessment. An end-to-end benchmark includes tasks with simple instructions, requiring step-by-step solutions to reach the final answers. A fine-grained benchmark contains tasks with a detailed assessment of essential skills in the web environment such as Optical Character Recognition (OCR), and semantic understanding.}
  \label{tab:web-bench}
\end{table*}

\setlength{\tabcolsep}{1pt}
\begin{table*}[!htb]
    \centering
    \footnotesize
    \renewcommand\tabcolsep{2pt}
    \renewcommand{\arraystretch}{1.0}
    \begin{tabular}{>{\arraybackslash}p{2.5cm} >{\centering\arraybackslash}p{2cm} >{\centering\arraybackslash}p{2.5cm} >{\centering\arraybackslash}p{2cm} >{\arraybackslash}p{6cm}} 
        \toprule 
        \textbf{Paper} & \textbf{Scene} & \textbf{Dataset} & \textbf{Method} & \textbf{Task} \\ 
        \midrule 
         \citet{li2023prompt} & Hotel, Movies \& TV, Restaurant & TripAdvisor, Amazon, Yelp & Embeddings,
         Prompting, Fine-tuning & Aspect extraction, Rating Prediction \\
        \midrule 
         P5 \cite{geng2022recommendation} & Sports, Beauty, Toys, Yelp & Amazon \cite{ni2019justifying}, Yelp & Pretraining, Prompting & Rating Prediction, Sequential Recommendation, Explanation Generation, Review Generation, and Direct Recommendation \\
        \midrule
        PETER \citet{li2021personalized} & Hotel, Movies \& TV, Restaurant & TripAdvisor, Amazon, Yelp & Transformer & Rating prediction and Explanation Generation \\
        \midrule
        PEPLER \cite{li2023personalized} & Hotel, Movies, TV and Restaurant & TripAdvisor5 (Hotel), Amazon (movies\& TV) and Yelp7 (restaurant) & Prompting, Fine-tuning & Explanation Generation\\
        \midrule 
        PALR \cite{yang2023palr} & Movies, Beauty & MovieLens-1M \cite{harper2015movielens}, Amazon Beauty \cite{ni2019justifying} & Fine-tuning, User Profile Generation, Retrieval & User Profile Generation and Direct Recommendation \\
        \midrule
        \citet{chu2023leveraging} & Sports, Outdoors, Beauty, Toys and Games & Amazon & Fine-tuning & Rating Prediction, Sequential Recommendation,  Direct Recommendation, Explanation Generation and Review Summarization \\
        \midrule
        \citet{liu2023chatgpt} & Beauty & Amazon & Prompting & Rating Prediction, Sequential Recommendation, Direct Recommendation, Explanation Generation and Review Summarization\\
        \midrule
        \citet{zhang2023recommendation} & Video Games& Amazon & Instruction tuning & Sequential Recommendation and Direct Recommendation \\
        \midrule
        \citet{hou2024large} & Movies & Amazon \cite{ni2019justifying}, MovieLens-1M \citet{harper2015movielens} &  Prompting  & Sequential Recommendation \\
        \midrule
        \citet{wang2023zeroshot} & Movies &  MovieLens-1M \cite{harper2015movielens} & Prompting & Sequential Recommendation and Direct Recommendation\\
        \midrule
        \citet{chen2022personalized} & News &  MIND \cite{wu-etal-2020-mind}, Reddit & Fine-tuning with weak labels & Direct Recommendation \\
        \bottomrule
    \end{tabular}
    \caption{An overview of existing research in recommendation. Following the classification of \citet{liu2023chatgpt}, we classify recommendation systems into five types: rating prediction, sequential recommendation, explanation Generation, and review generation,
    and direct recommendation.
    }
\label{tab:recommendation}
\end{table*}

\setlength{\tabcolsep}{1pt}
\begin{table*}[!htb]
    \centering
    \footnotesize
    \renewcommand\tabcolsep{2pt}
    \renewcommand{\arraystretch}{1.0}
    \begin{tabular}{>{\arraybackslash}p{5cm} >{\centering\arraybackslash}p{2cm} >{\centering\arraybackslash}p{2cm} >{\centering\arraybackslash}p{2cm} >{\centering\arraybackslash}p{2cm} > {\centering\arraybackslash}p{1.5cm} > {\centering\arraybackslash}p{1.5cm}}
        \toprule 
        \textbf{Dataset} &  \textbf{Scene} & \textbf{Task} & \textbf{\#Instances} & \textbf{\#Users} & \textbf{\#Items} \\ 
        \midrule 
        Amazon Review  \cite{ni2019justifying} & Products & Ratings, Reviews & 233.1M & 43.53M & 15.17M \\
        \midrule 
        MovieLens  \cite{harper2015movielens} & Movies & Ratings & 100,000 & 1,000 & 1,700 \\
        \midrule 
        Yelp  \cite{yelp} & Businesses & Ratings \& Reviews & 6,990,280 & 1,987,897 & 150,346 \\
        \midrule 
        TripAdvisor  \cite{li2023personalized}   & Hotels, Restaurants & Ratings \& Reviews & 320,023 & 9,765 & 6,280  \\
        \midrule 
        MIND  \cite{wu-etal-2020-mind} & News & Sequence recommendation & 15M & 1M & 160k \\
        \bottomrule
    \end{tabular}
    \caption{A list of commonly used datasets in personalized LLMs for recommendation and search task. 
    For the fifth column, the instances include reviews and ratings.}
    \label{tab:datasets_recommendation_search}
\end{table*}

\setlength{\tabcolsep}{1pt}
\begin{table*}[ht!]
    \centering
    \footnotesize
    \renewcommand\tabcolsep{2pt}
    \renewcommand{\arraystretch}{1.2}
    \begin{tabular}{>{\raggedright}m{3cm} l c c c r c} 
        \toprule 
        \textbf{Category} & \textbf{Dataset} & \textbf{\#Dialogues} & \textbf{\#Utterance} & \textbf{\#Domains} \\ 

        \midrule
        \multirow{8}{*}{{ToD}} 
        & MultiWOZ 1.0~\cite{budzianowski-etal-2018-multiwoz} & 10,438 & 75,894 & 7 \\
        & MultiWOZ 2.0~\cite{ramadan-etal-2018-large} & 8,438 & 63,841 & 7 \\
        & MultiWOZ 2.1~\cite{eric-etal-2020-multiwoz} & 7,032 & 57,022 & 7 \\
        & MultiWOZ 2.2~\cite{zang-etal-2020-multiwoz} & 10,438 & 71,572 & 7 \\
        & SGD~\cite{rastogi2020towards} & 22,825 & 463,284 & 20 \\
        & STAR~\cite{mosig2020star} & 6,652 & 127,833 & 13 \\
        & AirDialogue~\cite{wei-etal-2018-airdialogue} & 4,000 & 52,000 & 1 \\
        & UniDA~\cite{he2022galaxy} & 70,726 & 975,780 & 13 \\

        \midrule
        \multirow{6}{*}{{User Persona}} 
        & PersonaChat~\cite{zhang2018personalizing} & 11,907 & 164,356 & 1 \\
        & ConvAI2~\cite{dinan2019second} & 13,500 & 182,150 & 1 \\
        & Baidu PersonaChat~\cite{baidu-personachat} & 20,000 & 280,000 & 1 \\
        & JPersonaChat~\cite{sugiyama2021empirical} & 10,000 & 140,000 & 1 \\
        & JEmpatheticDialogues~\cite{sugiyama2021empirical} & 25,000 & 350,000 & 1 \\
        & DailyDialog~\cite{li-etal-2017-dailydialog} & 13,118 & 102,979 & 10 \\
        \bottomrule
    \end{tabular}
    \caption{A list of commonly used datasets for ToD modeling and user persona modeling. Among them, different versions of MultiWOZ~\cite{budzianowski-etal-2018-multiwoz,ramadan-etal-2018-large,eric-etal-2020-multiwoz,zang-etal-2020-multiwoz} and PersonaChat~\cite{zhang2018personalizing} are the most commonly used. Updated versions of MultiWOZ improve in several aspects: data quality, dialogue complexity, schema and ontology updates, and dataset sizes. PersonaChat contains various persona profiles, consisting of background, preferences, and personality traits. These profiles enable the modeling of coherent and contextual multi-turn diverse dialogue scenarios. For applications in user persona modeling, \citet{tu2023characterchat} match individuals with persona-compatible virtual supporters and introduces the MBTI-S2Conv dataset, containing conversations between characters with distinct profiles. \citet{lotfi2024personalitychat} and \citet{han2024psydial} both propose synthetic datasets related to the Big Five personality.}
    \label{tab:diag_data}
\end{table*}

\begin{table*}[htp!]
    \centering
    \footnotesize
    \renewcommand\tabcolsep{2pt}
    \renewcommand{\arraystretch}{1.2}
    \begin{tabularx}{\textwidth}{>{\raggedright}m{2.3cm}X}
        \toprule
        \textbf{Domain} & \textbf{Dialogue, Slots, and State} \\
        \midrule
        \multirow{5}{=}{\textbf{Restaurant}} & 
        [USER:] I need a place to dine in the center that's expensive.\\
        & \textbf{Slots:} restaurant-area: centre, restaurant-pricerange: expensive\\
        & \textbf{State:} active\_intent: find\_restaurant\\
        \cmidrule(l){2-2}
        & [SYSTEM:] I have several options for you; do you prefer African, Asian, or British food?\\
        & \textbf{State:} active\_intent: find\_restaurant, requested\_slots: restaurant-food\\
        \cmidrule(l){2-2}
        & [USER:] Any sort of food would be fine, as long as it is a bit expensive. Could I get the phone number for your recommendation?\\
        & \textbf{Slots:} restaurant-area: centre, restaurant-pricerange: expensive, restaurant-food\\
        & \textbf{State:} active\_intent: find\_restaurant, requested\_slots: restaurant-phone\\
        \cmidrule(l){2-2}
        & [SYSTEM:] There is an African place named Bedouin in the center. How does that sound?\\
        & \textbf{Slots:} restaurant-name: Bedouin\\
        \cmidrule(l){2-2}
        & [USER:] Sounds good, could I get that phone number? Also, could you recommend me an expensive hotel?\\
        & \textbf{Slots:} restaurant-area: centre, restaurant-name: Bedouin, restaurant-pricerange: expensive, restaurant-phone\\
        & \textbf{State:} active\_intent: find\_restaurant, requested\_slots: restaurant-phone\\
        \midrule

        \multirow{6}{=}{\textbf{Hotel}} & 
        [SYSTEM:] Bedouin's phone is 01223367660. As far as hotels go, I recommend the University Arms Hotel in the center of town.\\
        & \textbf{Slots:} hotel-name: University Arms Hotel\\
        & \textbf{State:} active\_intent: find\_hotel\\
        \cmidrule(l){2-2}
        & [USER:] Yes. Can you book it for me?\\
        & \textbf{State:} active\_intent: find\_hotel, requested\_slots: hotel-booking\\
        \cmidrule(l){2-2}
        & [SYSTEM:] Sure, when would you like that reservation?\\
        \cmidrule(l){2-2}
        & [USER:] I want to book it for 2 people and 2 nights starting from Saturday.\\
        & \textbf{Slots:} hotel-bookday: Saturday, hotel-bookpeople: 2, hotel-bookstay: 2\\
        & \textbf{State:} active\_intent: book\_hotel, slot\_values: hotel-bookday: Saturday, hotel-bookpeople: 2, hotel-bookstay: 2\\
        \cmidrule(l){2-2}
        & [SYSTEM:] Your booking was successful. Your reference number is FRGZWQL2. May I help you further?\\
        & \textbf{Slots:} hotel-bookday: Saturday, hotel-bookpeople: 2, hotel-bookstay: 2, reference number: FRGZWQL2\\
        \cmidrule(l){2-2}
        & [USER:] That is all I need to know. Thanks, goodbye.\\
        & [SYSTEM:] Thank you so much for Cambridge TownInfo center. Have a great day!\\
        \bottomrule
    \end{tabularx}
    \caption{An example of ToD modeling from the MultiWOZ dataset~\cite{zang-etal-2020-multiwoz}.}
    \label{tab:dialogue_example}
\end{table*}
\begin{table*}[htb!]
    \centering
    \footnotesize
    \renewcommand\tabcolsep{2pt}
    \renewcommand{\arraystretch}{1.2}
    \begin{tabularx}{\textwidth}{>{\raggedright}m{2.3cm}X}
        \toprule
        \textbf{Persona} & \textbf{Chat} \\
        \midrule
        \multirow{16}{=}{I fly airplanes.\\ I enjoy building computers.\\ My favorite band is tool.\\ I am in the army.\\ I dropped out of college.} &
        [PERSON 1:] Hello how are u tonight\\
        & [PERSON 2:] Hi. I am okay. tired, but okay. how are you ?\\
        & [PERSON 1:] I am doing good should be sleeping i have school but can't sleep \\
        & [PERSON 2:] I did not finish school, I enlisted in the army instead .\\
        & [PERSON 1:] Wow I am only 14 so I can't do that just yet but I hope too \\
        & [PERSON 2:] nice. stay in school and work hard . \\
        & [PERSON 1:] I try i like video games and race cars \\
        & [PERSON 2:] I like video games too, fallout is my favorite. \\
        & [PERSON 1:] I am a call of duty girl i can't wait for the new one \\
        & [PERSON 2:] My younger brother is a cod player too. he is pretty good .\\
        & [PERSON 1:] I have three best friends but lots of other friends that play it\\
        & [PERSON 2:] I have a best friend, she is a pilot like me. \\
        & [PERSON 1:] What kind of plane do u fly\\
        & [PERSON 2:] A bomber, it is awesome. do you want to take lessons \\
        & [PERSON 1:] I am kinda afraid of heights so not sure flying is for me\\
        & [PERSON 2:] You should at least try to go up in a plane, it is a blast. \\
        \bottomrule[1.5pt]
    \end{tabularx}
    \caption{An example of user persona modeling~\Cref{subsub:3-diag} from Persona-Chat dataset~\cite{zhang-etal-2018-personalizing}.}
    \label{tab:dialogue_persona}
\end{table*}

\end{document}